\documentclass[runningheads]{llncs}
\usepackage{graphicx}
\usepackage{bibnames}
\begin{document}
\title{Spatial Layout Consistency for 3D Semantic Segmentation}
\author{Maryam Jameela\inst{1}\and
Gunho Sohn\inst{1} }
\authorrunning{M.Jameela \& G.Sohn}
\institute{ \textsuperscript{1} Department of Earth and Space Science and Engineering\\
Lassonde School of Engineering, York University,\\
Toronto, ON M3J 1P3, Canada\\
\email{\{maryumja,gsohn\}@yorku.ca}}
\maketitle              % typeset the header of the contribution
\begin{abstract}
Due to the aged nature of much of the utility network infrastructure, developing a robust and trustworthy computer vision system capable of inspecting it with minimal human intervention has attracted considerable research attention. The airborne laser terrain mapping (ALTM) system quickly becomes the central data collection system among the numerous available sensors. Its ability to penetrate foliage with high-powered energy provides wide coverage and achieves survey-grade ranging accuracy. However, the post-data acquisition process for classifying the ALTM’s dense and irregular point clouds is a critical bottleneck that must be addressed to improve efficiency and accuracy. We introduce a novel deep convolutional neural network (DCNN) technique for achieving voxel-based semantic segmentation of the ALTM’s point clouds. The suggested deep learning method, Semantic Utility Network (SUNet) is a multi-dimensional and multi-resolution network. SUNet combines two networks: one classifies point clouds at multi-resolution with object categories in three dimensions and another predicts two-dimensional regional labels distinguishing corridor regions from non-corridors. A significant innovation of the SUNet is that it imposes spatial layout consistency on the outcomes of voxel-based and regional segmentation results. The proposed multi-dimensional DCNN combines hierarchical context for spatial layout embedding with a coarse-to-fine strategy. We conducted a comprehensive ablation study to test SUNet’s performance using  67  km  x  67  km  of  utility  corridor  data  at  a  density of ${5pp/m^2}$. Our experiments demonstrated that SUNet’s spatial layout consistency and a multi-resolution feature aggregation could significantly improve performance, outperforming the SOTA baseline network and achieving a good F1 score for pylon  (89\%), ground (99\%), vegetation (99\%) and powerline (98\%) classes.

\end{abstract}

\sloppy
\section{Introduction}\label{Introduction}
 
% KAO: Sloppy spacing ensures non-overfull lines. Can be removed if this is not an issue.
\sloppy
Conducting a safe, efficient, and accurate inspection for utility network management is vital to grid stability and resilience for protecting our economy. Traditionally, the utility network inspection is performed by a ground crew to manage vegetation encroachment or monitor the physical condition of the utility infrastructure \cite{3dclassRF}. Recently, operating an unmanned aerial vehicle (UAV) for utility inspection is gaining popularity to reduce the cost of data collection \cite{isprs-annals-IV-2-W7-227-2019} \cite{rtpowerline}. However, it is still an open challenge to operate UAVs for completing data collection covering entire utility corridors due to strict flying regulations, short operation time, and limited spatial coverage. Thus, ALTM has become a central data acquisition platform for many utility network inspection projects. ALTM acquires dense point clouds to represent the utility infrastructure and its surrounding environment with survey-grade ranging accuracy \cite{randmforestPC}. A high-powered laser beam can penetrate foliage, enabling three-dimensional (3D) vegetation encroachment analysis at a single tree level \cite{pl3dRecons}\cite{SVMPowerline}. However, performing visual perception tasks such as labeling point clouds with semantic attributes from ALTM point clouds is still costly and conducted by laborious and error-prone manual work. Thus, there is an urgent demand to automate the post-data acquisition process to minimize human intervention. 
\begin{figure}[!t]
\centering
\includegraphics[width=4in]{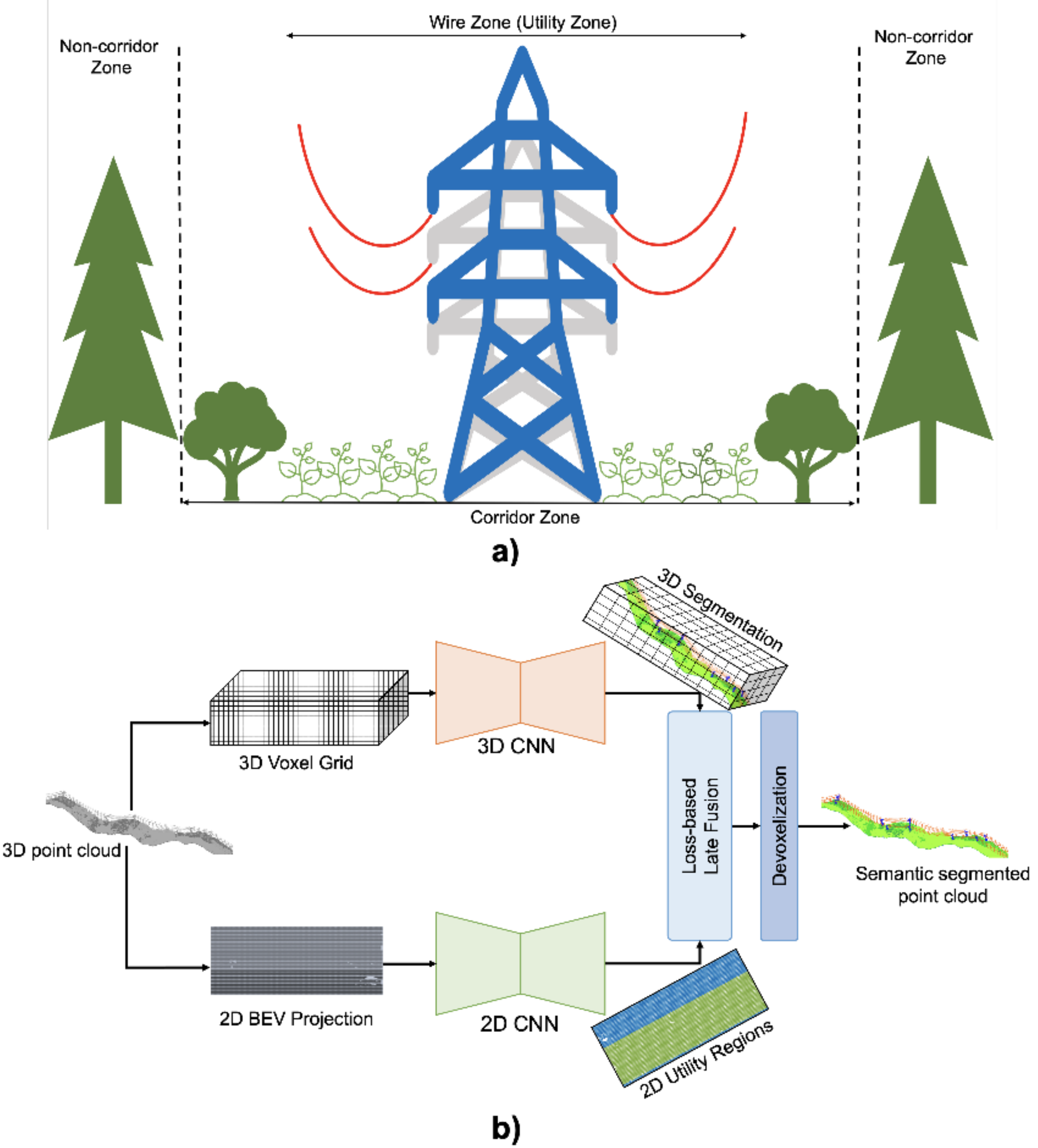}
\caption{SUNet is a multi-dimensional and multi-resolution network which imposes the spatial layout consistency (1a) through 2D bird's-eye view (BEV) of utility regions on the outcomes of 3D segmentation network via loss-based late fusion (1b).}
\label{fig:fig1}
\end{figure}
Recently, DCNN has outperformed previous state-of-the-art (SOTA) machine learning techniques with large margins in computer vision \cite{kpconv}. DCNN’s most significant advantage is learning high-level features from data in an incremental manner. This deep feature representation generally generalizes DCNN’s performance easier than other machine learning techniques. Recently several research groups have proposed the successful design of deep neural networks for achieving semantic segmentation using point clouds \cite{pointnet,pointnetplusplus,kpconv}. Though, neither have utilized the spatial layout found through the infrastructures, especially for utility corridors nor embedded the spatial layout for global context. This limitation inspired us to propose a network that deals with hierarchical spatial consistency that can later be generalized for any standard layout segmentation problem.\\ 
This research thoroughly investigates the utility corridor dataset.  Firstly, the study unraveled spatial layout consistency. It identifies the hierarchy between regions (corridor and non-corridor) and object classes such as ground, towers, powerlines, and vegetation as shown in figure \ref{fig:fig1}.  Secondly, it utilizes the hierarchical spatial layout consistency by combining multi-dimensional input. Finally, it explores the science of human vision and information extraction for decision-making as a hierarchy.  We proposed SUNet which combines two networks: one classifies point clouds at multi-resolution with object categories in three dimensions and another predicts two-dimensional regional labels distinguishing corridor regions from non-corridors. A significant innovation of the SUNet is that it imposes spatial layout consistency on the outcomes of voxel-based and regional segmentation results. SUNet also has a multi-resolution feature aggregation module in a three-dimensional network for enhancing the receptive field for minority class prediction.
The following sections will discuss related work, proposed methodology, experiments, and results.

\section{Literature Review}\label{sec:Literature Review}
We will discuss various groups of semantic segmentation methods for 3D point clouds and the importance of spatial layout in this section.
\subsection{Semantic Segmentation}
Semantic segmentation for 3D point cloud has improved significantly in the last decade, especially after the mainstream use of deep learning for computer vision. Most of the research focused on utilizing intrinsic, extrinsic, and deep features for semantically labeling each point with an enclosing object or region \cite{randla}.
One of the major limitations observed is the lack of focus on imposing spatial layout consistency from the real world for segmentation and global context embedding. Existing methods can be divided into three sections, i) statistical segmentation, ii) machine learning-based classification networks and iii) deep learning-based segmentation networks.  
\subsubsection{Geometrical Segmentation} Multiple existing methods treat utility corridor segmentation as a geometrical problem. These methods extract lines, cluster point clouds, and classify them using neighborhood votes, density, and height-based features \cite{HTStatistical}. These methods have shown tremendous improvement over time. The major limitations of these techniques are preprocessing, feature crafting, and the requirement of extensive domain knowledge. Most of these methods require multiple steps of filtering to segment the regions \cite{automatedPowerlineExtraction}. These techniques are not robust for raw large-scale dense 3D point clouds \cite{pl3dRecons}.
\subsubsection{Machine Learning based Utility Classification} Previous literature dealing with utility object classification, powerline reconstruction, and extraction has shown exemplary performance \cite{MLSpan3DPL}. These methods employ support vector machines \cite{SVMPowerline}, random forest \cite{3dclassRF,randmforestPC}, decision trees, and balanced/unbalanced learning. These systems use either 3D voxels or 2D grid projections and handcraft features from eigenvalues to facilitate decision-making by providing distinguishing information \cite{rtpowerline}.  Most of these techniques have shown limitations on 3D large-scale datasets. Deep learning and computer vision provided the ability to encode features that helps in decoding information for almost every computer vision task, including segmentation and classification. It liberated researchers from possessing extensive domain knowledge required for crafting features. It also generalizes the solutions across datasets and sensors.
\subsubsection{Deep Learning-based Segmentation Networks}  Deep learning research has provided academic and commercial communities with a strong foundation for integrating artificial intelligence in post-processing, analysis, and predictive models. Semantic segmentation of point cloud has shown high performance using 3D voxels and 2D bird-eye view. These representations inherently come with the limitation of quantization errors despite providing an efficient boost to performance. The new batch of methods pioneered by PointNet \cite{pointnet} began to use raw point cloud directly as input to deep learning network which was extended by PointNet++ \cite{pointnetplusplus}, KPConv introduced continuous kernels \cite{kpconv} and RandLA  is state of the art network\cite{randla}. These networks have shown impressive and comparable performance on most segmentation benchmarks; such as Semantic3D \cite{semantic3d}, SensatUrban \cite{urbansense}, or DublinCity \cite{dublincity}. Though, neither have utilized the spatial regularity found through the infrastructures, especially for utility corridors nor embedded the spatial layout for global context.  This limitation inspired us to propose a SUNet.
\subsection{Spatial Layout}
Several studies in cognitive science, architectural design, and civil engineering utilize the spatial layout and relationships of objects with respect to each other \cite{archtech}. This relationship helps our brains to understand the context of the scene and interpret it for decision-making. Convolution neural networks were designed to extract this spatial relationship but for learning global context spatial regularities needed to be embedded in the network such as railway lane extractions, road lane detection \cite{railwayextraction} and 3D model of the building \cite{cnnspatial}. Various other studies have been able to demonstrate how small objects such as cars on the side of a road will be easily detected based on the spatial relationship which might otherwise be ignored \cite{image2Object}. Our study is inspired by this research to use the spatial relationship between corridor non-corridor regions and objects present in these regions to demonstrate the importance of spatial layout embedding.

\section{Methodology}\label{sec:methodology}
SUNet is a multi-dimensional and multi-resolution network. We propose to decode the spatial consistency between layout and objects of interest. It consists of two networks; two-dimensional regional prediction network \cite{2DUnet}, which constrains the predictions of another three-dimensional network through loss-based late fusion.
\begin{figure*}
\centering
\includegraphics[width=13cm]{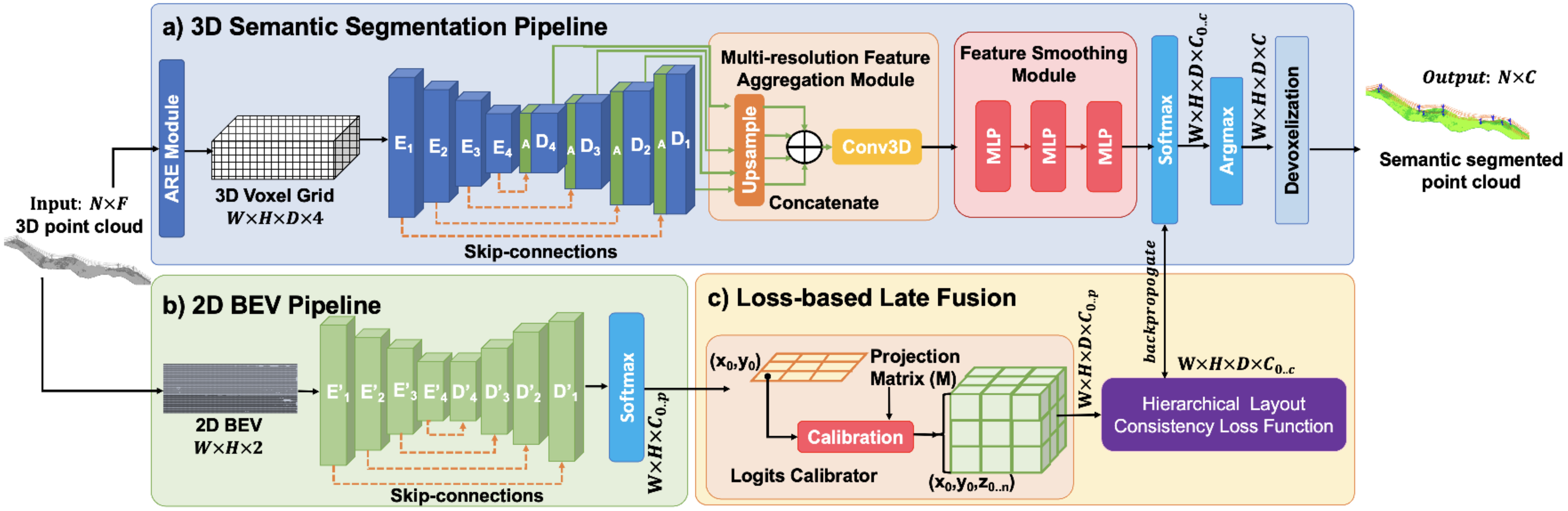}
  \caption{The overall architecture of SUNet: Point cloud is preprocessed into voxel grid and BEV. These are separately processed by a multi-resolution 3D semantic segmentation pipeline and 2D BEV pipeline for regional prediction to impose the spatial layout consistency on 3D objects through loss-based late fusion.}
  \label{fig:fignetwork}
\end{figure*}
\subsection{3D Semantic Segmentation Pipeline}
Overall network architecture is illustrated in figure \ref{fig:fignetwork}. 3D segmentation network is a U-shaped multiresolution encoder-decoder \cite{3DUNET,m3DUnet}. It extracts features from a three-dimensional voxel grid to predict semantic labels for object classes \cite{att3DUnet}. In particular, this network extracts four feature maps ${E_1}$, ${E_2}$, ${E_3}$, and ${E_4}$. As shown in figure \ref{fig:fignetwork} ${E_n}$ constructs output feature map ${D_n}$ with series of operations including additive attention applied on ${D_{n-1}}$ and ${E_n}$, two 3x3x3 convolution operations followed by batch normalization and relu activation. It results in four output feature maps \{ ${D_1}$, ${D_2}$, ${D_3}$,${D_4}$ \} with dimensions ${\frac{H}{2l}\times\frac{W}{2l}\times\frac{D}{2l}\times {32l}}$ for output D at level ${l}$. These feature maps represent the deep multi-resolution output which in terms of semantic segmentation provides a multi-receptive field for segmenting objects and areas of different sizes. These feature maps are then passed through a multi-resolution feature aggregation module to incorporate knowledge and context from all levels. The final part of the pipeline is the feature smoothing module which eliminates noisy and cluttered predictions and delivers the confidence score against each class. This is passed by a loss-based late fusion module to refine and constrain using spatial layout consistency and back-propagate the loss to better learning deep features. 
\subsubsection{Additive Attention Module}
 In multi-resolution encoder-decoder architecture, downsampling can result in an overpopulation of false positives for small objects with significant shape variance. An additive attention gate can fix those localization problems and reduce misclassification \cite{att3DUnet}. This module learns multiple attention co-efficient for each class and produces the activated feature maps pruned to suppress the irrelevant information. These attention gates take input from previous layer of decoder ${D_1}$ for ${D_2}$ and encoder feature map of same level ${E_2}$ and generate attention co-efficient ${\alpha}$ to take element-wise summation with feature map ${E_2}$. 
\subsubsection{Multi-resolution Feature Aggregation Module}
 Our proposed module aggregates the feature maps from \{ ${D_1}$, ${D_2}$, ${D_3}$,${D_4}$ \} by upsampling and concatenating. Aggregated feature map then passes through 1x1x1 convolution. It helps in generating a feature map that eliminated the error introduced by upsampling of the decoder for minority class prediction.
\subsubsection{Feature Smoothing Module} 
It is an optional module that has three multi-layer perceptron (MLP) operations to smooth the feature maps.  It evenly distributes the feature mean and variances. We plugged this module to remove the misclassification by a minimum margin. 
\subsection{2D BEV Pipeline}
The biggest challenge for semantic segmentation is to integrate a larger receptive field and global context into the network for spatially informed prediction. Humans’ judgment of scene semantics largely relied on their ability to grasp the global context. Our three-dimensional segmentation network is limited to local context and lacks coarser details of the scene for encoding spatial layout and global correlation of the objects.  Our 2D BEV pipeline fuse this information through loss based late fusion module. Two-dimensional network is a multi-resolution encoder-decoder shown in figure \ref{fig:fignetwork} as a regional semantics block \cite{2DUnet}. It consists of the four feature maps \{${E'_1}$, ${E'_2}$, ${E'_3}$,${E'_4}$\} which constructs four output maps \{${D'_1}$, ${D'_2}$, ${D'_3}$,${D'_4}$\}. This network takes BEV of a complete 3D point cloud scene and predicts regional classes probability tensor of shape ${W\times H \times C_p}$.
\subsection{Loss based Late Fusion} 
This is a key module that is further divided into two sub-modules. The following sections will discuss this in detail.
\subsubsection{Logits Calibrator}
This module is straightforward calibration that takes the projection matrix between 3D voxels and 2D BEV and converts the 2D logits for regional classes of shape ${W\times H \times C_p}$ to 3D representation ${W\times H \times D \times C_p}$ by exploring the one-to-many relationship.
\subsubsection{Hierarchical Layout Consistency (HLC) Loss Function}
The proposed loss function takes advantage of hierarchy by penalizing the misclassification of object classes in the incorrect regions (corridor and non-corridor). Our loss function is a key element of late fusion module for fusing the context from a local and global scale. It takes 3D predictions from 3D pipeline and refines them through a global spatial layout extracted from 2D pipeline to bridge the gap of the larger receptive field. It evaluates the target class correlation $y_{m}^{P_c}$ to penalize the prediction of $x_m$ weighted by $w_p$,$w_c$ to reduce loss for better prediction as shown in equation \ref{equ:1}. These co-occurrence weights for our experiments was 10.0 and 8.0 respectively and could be tuned. $P$ is total number of parent classes which is two and $C$ is total number of child classes.

\begin{equation}\label{equ:1}
	L=\frac{1}{M}\sum_{p=1}^{P}\sum_{c=1}^{C} \sum_{m=1}^{M} (w_p w_c)\times (y_{m}^{P_c} \times (log(h_{\Theta}(x_m, c, p))))
\end{equation}
\subsection{Voxelization and BEV Projection}
Input representation of the 3D point cloud for our segmentation network is a voxel grid. We preprocess the raw point cloud onto the voxel grid and calculate a mean representing all the points residing in the 3D voxel. The network maintains a projection matrix from the voxel grid to the raw point cloud for easy label projection. This voxel grid provides a trade-off between efficiency and effectiveness based on the selection of voxel size.\\
BEV is a 2D view of a 3D point cloud. Our 2D BEV pipeline takes XY projection of a 3D scene where each pixel represents the residing points. XY-projection gave the most optimal BEV for extracting global context for regional and object prediction. Our projection matrix M between BEV and 3D voxel grid provides projection compatibility between feature spaces to fuse 2D and 3D predictions for better utilization of spatial layout consistency. 
\subsection{Absolute and Relative Elevation Module}
 The data analysis showed non-homogeneous areas where points belonging to the same classes could be on different elevations. Hence, we incorporate local pillar-based normalization from equation \ref{equ:4} which uses the min elevation of each column and the maximum elevation of the scene for normalization and calculates over the whole point cloud $\{i\ni N, N =total \, points\}$.\\
\textbf{Global Normalization:}
\begin{equation}\label{equ:3}
Absolute \, Elevation_i = 1+\frac{z_i}{z_{max\_global}}\
\end{equation}
\textbf{Local Normalization:}
\begin{equation}\label{equ:4}
Relative \, Elevation_i= \frac{(z_i-z_{min\_local})}{(z_{max\_scene}-z_{min\_local})}\
\end{equation}

\begin{table}
\renewcommand{\arraystretch}{1.3}
\caption{Ablation Study on importance of feature engineering over F1-Score; Occ: No. of Occupancy Points, AE: Absolute Elevation, RE: Relative Elevation and NR: Number of Returns }
	\centering
		\begin{tabular}{p{2.5cm}|p{3.5cm}|p{1.2cm}|p{1.5cm}|p{1.0cm}|p{1.5cm}}\hline
			{Methods}&{Features} & {Pylon (\%)} &{Ground (\%)} &{Veg (\%) }&{Powerline (\%) }\\\hline
			 Baseline&Occ + AE + NR  & 77.0 & 93.0& 97.0 & 91.0\\\hline
			 SUNet \textbf{(ours)}  & Occ + AE + NR +RE & \textbf{84.0} &\textbf{99.0}&\textbf{99.0} &	\textbf{98.0}\\\hline
		\end{tabular}
	\label{tab:trainfe}
\end{table}

\begin{table*}
\renewcommand{\arraystretch}{1.3}
\caption{Comparison of baseline 3D Attention UNet (3D AUNet) to proposed SUNet and ablation study of modules  multiresolution feature aggregation (MFA), and feature smoothing (FS)  on recall (R), precision (P) and F1 score (F1).}
	\centering
		\begin{tabular}{p{2.6cm}|p{0.7cm}p{0.7cm}p{0.7cm}|p{0.8cm}p{0.8cm}p{0.8cm}|p{0.7cm}p{0.7cm}p{0.7cm}|p{0.7cm}p{0.7cm}p{0.7cm}}\hline
			{Methods} & \multicolumn{3}{|c}{Pylon (\%) } & \multicolumn{3}{|c}{Ground (\%)} &\multicolumn{3}{|c}{Veg (\%) }&\multicolumn{3}{|c}{Powerline (\%)}\\\hline
			  & R & P & F1& R & P & F1& R & P & F1& R & P & F1\\\hline
			 3D AUNet (baseline)  & 72.0 &	84.0& 77.0 & 93.0 & 94.0& 93.0& 97.0 &97.0 &97.0 &	84.0&98.0&91.0\\\hline
			 SUNet+MFA+FS \textbf{(ours)} 	& 78.0 &97.0&87.0&	\textbf{100}&99.0&\textbf{100}& 99.0 &99.0&99.0&	\textbf{99.0}&97.0&98.0\\\hline
			 \textbf{SUNet+MFA \textbf{(ours)}} 	& \textbf{82.0} & \textbf{96.0} & \textbf{89.0} &	99.0 & \textbf{99.0} & 99.0 &	\textbf{99.0} & \textbf{99.0} & \textbf{99.0}&	98.0& \textbf{99.0} & \textbf{99.0} \\\hline
			 
		\end{tabular}
	\label{tab:train}
\end{table*}
\sloppy
\begin{figure*}
\centering
\includegraphics[width=1\linewidth]{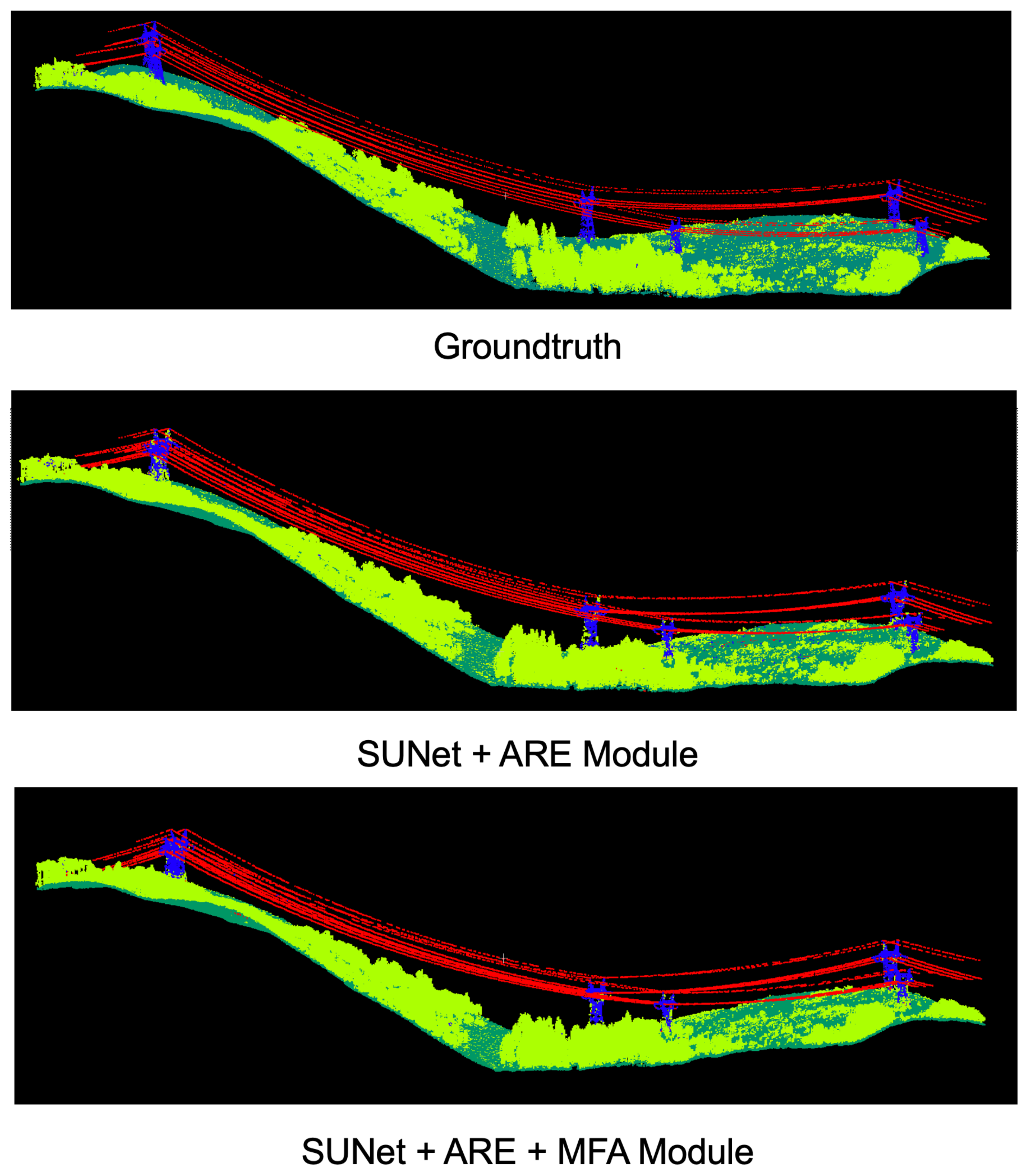} 
\caption{Visualization of results; Groundtruth, SUNet+ ARE (ours) and SUNet + ARE + Multiresolution feature aggregation module (ours). Blue: pylon, red: powerline, green: high vegetation and dark-green: ground }
\label{fig:figSideView}
\end{figure*}

\begin{figure*}
\centering
\includegraphics[width=1\linewidth]{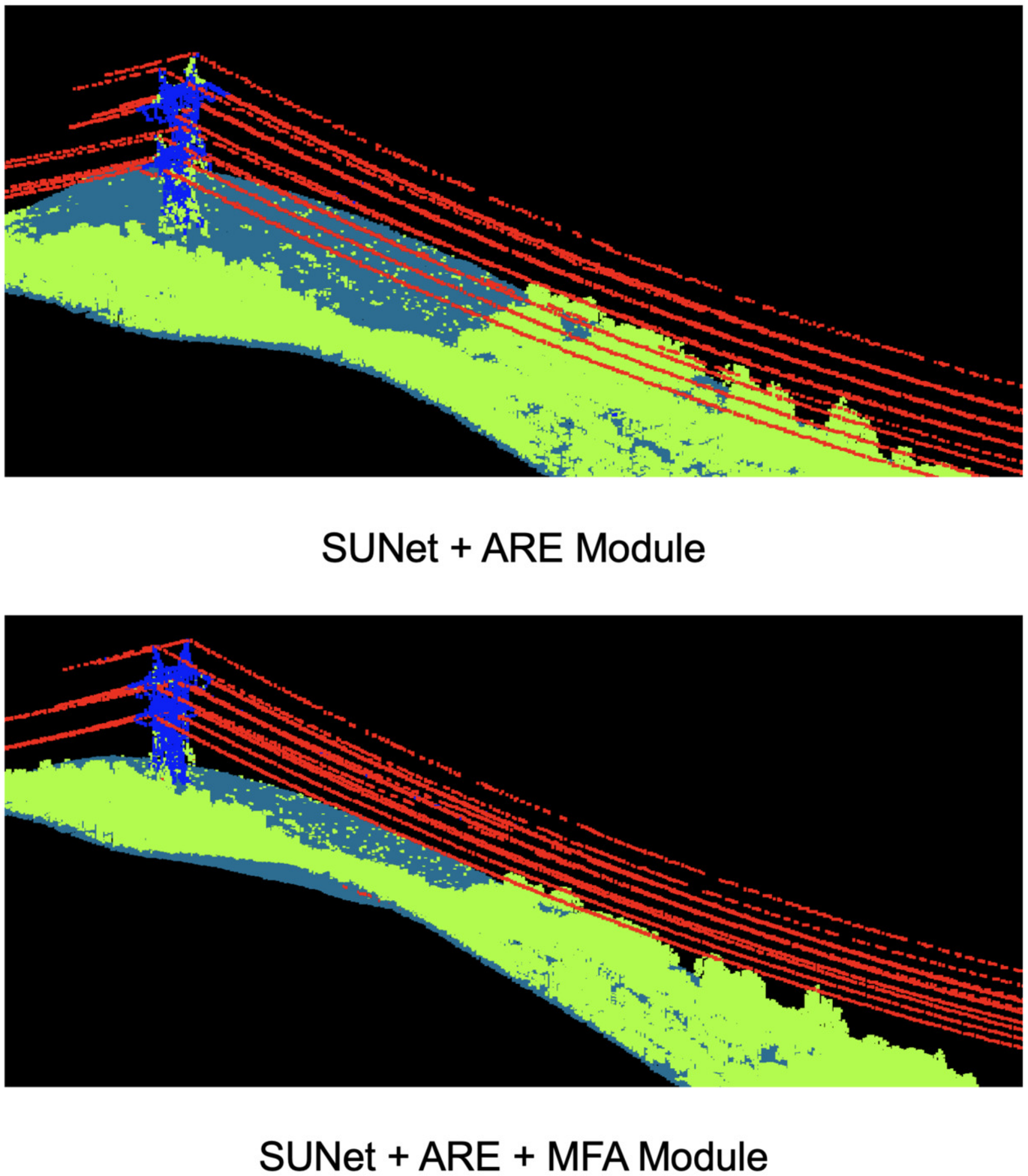} 
\caption{\label{tab:results}Visualization of Errors;  SUNet+ ARE (ours) and SUNet + ARE + Multiresolution feature aggregation module (ours). Blue: pylon, red: powerline, green: high vegetation and dark-green: ground }
\label{fig:figError}
\end{figure*}

\section{Experiments and Results}\label{experiment}
We conducted ablation studies based on feature engineering, loss and proposed modules.  Experiments  assess test set on the ground, pylon, powerline, and vegetation classes. These four classes are important for the utility industry for predictive maintenance.
\subsection{Dataset}
 Data is collected using a Riegl Q560 laser scanner over  ${67km^2}$ in Steamboat Springs, Colorado, USA. Data is later divided into train and test sets for experiments. The first ${8km^2}$ of the dataset is used for testing and the remaining data is used for training the network. There is a total of 67 non-overlapping scenes each containing more than two million points with an average density of ${5pp/m^2}$. We manually labeled our ground truth using Terrasolid point cloud processing software. Labels were generated through extensive domain knowledge and technical expertise. Our training dataset had five classes powerline, pylon, low vegetation, ground, and medium-high vegetation. We merged low vegetation into ground class as the most ground is covered by low vegetation. We also labeled our regional classes based on utility community literature. Powerlines and pylons are in the corridor region and (3-10) m away from the pylon is a non-corridor region.
 \subsection{Experimental Configuration}
 We took half of the scenes to pretrain our 2D regional prediction network for global regional prediction of spatial layout. Each scene is divided into four subscenes by GPS time of flight line. These subscenes are then projected on a ${640\times 640}$ 2D BEV grid of each pixel size= ${1m^2}$. Each grid has two feature channels, mean elevation, and the number of occupancy points per pixel. The input size of the 2D network is ${640\times640\times2}$ and the output is ${640\times640\times3}$ which is the regional classes’ confidence score.
 We pretrained the network with ${batch size=1}$ and total epochs of 100 with K-cross validation to avoid overfitting. We augmented our dataset using horizontal flip, vertical flip and random rotation. The network is trained on two GPU RTX 6000; training time is between 4-5 hours, and inference takes about 30 seconds. 
 Our voxel grid is generated over each subscene and the size is ${640\times640\times448}$ with a voxel size of ${1m^3}$. Each batch consists of maximum elevation of the whole scene to provide network the complete view to better deal with vertical context. Feature channels consist of absolute and relative elevation as discussed in methodology, number of occupancy points, and a number of returns. We have conducted a feature engineering study to select these features which will be discussed in the following section.  SUNet outputs ${32\times32\times448\times5}$ which is a confidence score against 3D classes (background, pylon, powerline, vegetation, and ground). Later, the final prediction assigns a true label based on the highest confidence score and project voxels labels on points using the project matrix. SUNet is trained on two GPU RTX 6000 for 100 epochs; training time is between 48-60 hours, and inference takes about 2-3 mins. 
\subsection{Evaluation Metrics}
We selected precision, recall, and F1 score for evaluating the performance of our network. The precision determines the total number of predicted samples out of the total instances of the class. Recall measures correctly predicted samples out of the total predicted sample of the class.  F1 score shows the balance between precision and recall.

\subsection{Feature Engineering}
We progressively developed a feature engineering study and used a number of occupancy points and absolute elevation and number of returns as input. Our feature engineering study was also inspired by \cite{randmforestPC} to develop absolute and relative elevation modules to deal with non-homogeneous surfaces. SUNet with ARE module shows a drastic improvement in the powerline, pylon, and ground classes as shown in table \ref{tab:trainfe}. 
\begin{table}
\renewcommand{\arraystretch}{1.3}
\caption{Ablation Study on importance of loss function over F1-Score; WCE: Weighted Cross Entropy and HLC: Hierarchical Layout Consistency Loss}
	\centering
		\begin{tabular}{p{2.0cm}|p{2.3cm}|p{1.0cm}|p{1.5cm}|p{1.2cm}|p{1.5cm}}\hline
			{Methods}&{Loss Function} & {Pylon} &{Ground} &{Veg}&{Powerline}\\\hline
			 Baseline & WCE & 82.0& 94.0 &97.0 &91.0\\\hline
			 SUNet \textbf{(ours)}  &HLC& \textbf{84.0} &\textbf{99.0}&\textbf{99.0} &	\textbf{98.0}\\\hline
		\end{tabular}
	\label{tab:trainloss}
\end{table}
\subsection{Loss Functions}
We proposed HLC loss function to impose the spatial layout consistency over the 3D object segmentation. As shown in table \ref{tab:trainloss} the remarkable increase in the performance of all classes especially pylon and ground classes. 
\subsection{Module based Ablation Study}
Later, we had dived into a module-based ablation study. We decided on four features and plugged different combinations of modules to observe the performance. SUNet used attention UNet \cite{att3DUnet} as the backbone showing improvement in all classes, especially the pylon and powerline classes. This validates the significance of global context from the 2D network to refine predictions based on regional spatial layout embedding. We then introduced multi-resolution feature aggregation and feature smoothing modules. We aggregated features from all scales of the backbone network and that improved the precision of the pylon to 97\%. It shows that the pylon which is the smaller object and minority class is ignored by the downsampling. Though feature smoothing didn’t show any large impact on recall for pylon class. All other classes are already reaching (98-99)\%. We removed the feature smoothing module and recall for pylon class 82\% improved which showed that feature smoothing was causing an error due to the structural similarity of tree and pylon.
\subsection{Error Analysis}
The 2D regional prediction network has room for improvement. It is the reason that the pylon class is still below 90\% in the recall. Feature smoothing improved the precision of the pylon class as shown in table \ref{tab:train}. Though, Lower parts of pylons are misclassified as trees and isolated points labeled as powerlines. These errors are due to structural similarity of tree and pylon in terms of point density and data distribution as shown in figure \ref{fig:figError}.

\section{Conclusion}\label{conclusion}
In this work, we demonstrated that proposed SUNet with loss based late fusion contributes to embedding spatial layout. It also proved multi-resolution feature aggregation module incorporate the broader context for decision making. These experiments have verified conceptual inspiration and hypothesis i.e.; hierarchical layout spatial consistency combined with coarse-to-fine strategy can facilitate a development of deep semantic segmentation model for predictive analysis. Our future work will focus on improving the performance of regional predictions and generalization of the proposed network on various sensors.

\section{Acknowledgment}
This research project has been supported by the Natural Sciences and Engineering Research Council of Canada (NSERC)'s Collaborative Research and Development Grant (CRD) – 3D Mobility Mapping Artificial Intelligence (3DMMAI) and Teledyne Geospatial Inc. We'd like to thank Leihan Chen (Research Scientist), Andrew Sit (Product Manager), Burns Forster (Innovation Manager) and Chris Verheggen (SVP R\& D). 

% \bibliographystyle{end}

\bibliographystyle{splncs04}
\bibliography{output}

\begin{thebibliography}{10}
\providecommand{\url}[1]{\texttt{#1}}
\providecommand{\urlprefix}{URL }
\providecommand{\doi}[1]{https://doi.org/#1}

\bibitem{archtech}
Brosamle, M., Holscher, C.: Architects seeing through the eyes of building
  users, a qualitative analysis of design cases. 2007, International Conference
  on Spatial Information Theory (COSIT'07) pp. 8--13 (01 2007)

\bibitem{3DUNET}
{\c{C}}i{\c{c}}ek, {\"{O}}., Abdulkadir, A., Lienkamp, S.S., Brox, T.,
  Ronneberger, O.: 3d u-net: Learning dense volumetric segmentation from sparse
  annotation. International Conference on Medical Image Computing and
  Computer-Assisted Intervention (MICCAI,2016)  \textbf{9901},  424--432
  (2016). \doi{https://doi.org/10.1007/978-3-319-46723-849}

\bibitem{powerlineVeh}
Guan, H., Yu, Y., Li, J., Ji, Z., Zhang, Q.: Extraction of power-transmission
  lines from vehicle-borne lidar data. International Journal of Remote Sensing
  \textbf{37}(1),  229--247 (2016). \doi{10.1080/01431161.2015.1125549},
  \url{https://doi.org/10.1080/01431161.2015.1125549}

\bibitem{semantic3d}
Hackel, T., Savinov, N., Ladicky, L., Wegner, J.D., Schindler, K., Pollefeys,
  M.: Semantic3d.net: {A} new large-scale point cloud classification benchmark.
  CoRR  \textbf{abs/1704.03847} (2017), \url{http://arxiv.org/abs/1704.03847}

\bibitem{cnnspatial}
Haldekar, M., Ganesan, A., Oates, T.: Identifying spatial relations in images
  using convolutional neural networks. In: 2017 International Joint Conference
  on Neural Networks (IJCNN). pp. 3593--3600 (2017).
  \doi{10.1109/IJCNN.2017.7966308}

\bibitem{urbansense}
Hu, Q., Yang, B., Khalid, S., Xiao, W., Trigoni, N., Markham, A.: Towards
  semantic segmentation of urban-scale 3d point clouds: {A} dataset, benchmarks
  and challenges. CoRR  \textbf{abs/2009.03137} (2020),
  \url{https://arxiv.org/abs/2009.03137}

\bibitem{randla}
Hu, Q., Yang, B., Xie, L., Rosa, S., Guo, Y., Wang, Z., Trigoni, N., Markham,
  A.: Randla-net: Efficient semantic segmentation of large-scale point clouds.
  CoRR  \textbf{abs/1911.11236} (2019), \url{http://arxiv.org/abs/1911.11236}

\bibitem{m3DUnet}
Isensee, F., Kickingereder, P., Wick, W., Bendszus, M., Maier{-}Hein, K.H.:
  Brain tumor segmentation and radiomics survival prediction: Contribution to
  the {BRATS} 2017 challenge. CoRR  \textbf{abs/1802.10508} (2018),
  \url{http://arxiv.org/abs/1802.10508}

\bibitem{railwayextraction}
Jeon, W.G., Kim, E.M.: Automated reconstruction of railroad rail using
  helicopter-borne light detection and ranging in a train station. Sensors and
  Materials  \textbf{31}, ~3289 (10 2019). \doi{10.18494/SAM.2019.2433}

\bibitem{automatedPowerlineExtraction}
Jung, J., Che, E., Olsen, M.J., Shafer, K.C.: Automated and efficient powerline
  extraction from laser scanning data using a voxel-based subsampling with
  hierarchical approach. ISPRS Journal of Photogrammetry and Remote Sensing
  \textbf{163},  343--361 (2020).
  \doi{https://doi.org/10.1016/j.isprsjprs.2020.03.018}

\bibitem{MLSpan3DPL}
Jwa, Y., Sohn, G.: A multi-level span analysis for improving 3d power-line
  reconstruction performance using airborne laser scanning data. ISPRS -
  International Archives of the Photogrammetry, Remote Sensing and Spatial
  Information Sciences  \textbf{38},  97--102 (09 2010)

\bibitem{pl3dRecons}
Jwa, Y., Sohn, G., Kim, H.: Automatic 3d powerline reconstruction using
  airborne lidar data. IAPRS  \textbf{38},  105--110 (01 2009)

\bibitem{3dclassRF}
Kim, H., Sohn, G.: 3d classification of powerline scene from airborne laser
  scanning data using random forests. IAPRS  \textbf{38},  126--132 (09 2010).
  \doi{10.13140/2.1.1757.4409}

\bibitem{RFMClass}
Kim, H., Sohn, G.: Random forests based multiple classifier system for
  power-line scene classification. vol. XXXVIII-5/W12 (08 2011).
  \doi{10.5194/isprsarchives-XXXVIII-5-W12-253-2011}

\bibitem{randmforestPC}
Kim, H., Sohn, G.: Point-based classification of power line corridor scene
  using random forests. Photogrammetric Engineering and Remote Sensing
  \textbf{79},  821--833 (09 2013). \doi{10.14358/PERS.79.9.821}

\bibitem{HTStatistical}
Liu, Y., Li, Z., Hayward, R., Walker, R., Jin, H.: Classification of airborne
  lidar intensity data using statistical analysis and hough transform with
  application to power line corridors. In: 2009 Digital Image Computing:
  Techniques and Applications. pp. 462--467 (2009). \doi{10.1109/DICTA.2009.83}

\bibitem{3a226a67598044949605c22149f55a45}
Mottaghi, R., Chen, X., Liu, X., Cho, N., Lee, S., Fidler, S., Urtasun, R.,
  Yuille, A.: The role of context for object detection and semantic
  segmentation in the wild. In: Proceedings of the IEEE Computer Society
  Conference on Computer Vision and Pattern Recognition. pp. 891--898.
  Proceedings of the IEEE Computer Society Conference on Computer Vision and
  Pattern Recognition, IEEE Computer Society (Sep 2014).
  \doi{10.1109/CVPR.2014.119}, publisher Copyright: {\textcopyright} 2014
  IEEE.; 27th IEEE Conference on Computer Vision and Pattern Recognition, CVPR
  2014 ; Conference date: 23-06-2014 Through 28-06-2014

\bibitem{NAN2021212}
Nan, Z., Peng, J., Jiang, J., Chen, H., Yang, B., Xin, J., Zheng, N.: A joint
  object detection and semantic segmentation model with cross-attention and
  inner-attention mechanisms. Neurocomputing  \textbf{463},  212--225 (2021).
  \doi{https://doi.org/10.1016/j.neucom.2021.08.031},
  \url{https://www.sciencedirect.com/science/article/pii/S0925231221012157}

\bibitem{att3DUnet}
Oktay, O., Schlemper, J., Folgoc, L.L., Lee, M.C.H., Heinrich, M.P., Misawa,
  K., Mori, K., McDonagh, S.G., Hammerla, N.Y., Kainz, B., Glocker, B.,
  Rueckert, D.: Attention u-net: Learning where to look for the pancreas. CoRR
  \textbf{abs/1804.03999} (2018), \url{http://arxiv.org/abs/1804.03999}

\bibitem{9206883}
Peng, J., Nan, Z., Xu, L., Xin, J., Zheng, N.: A deep model for joint object
  detection and semantic segmentation in traffic scenes. In: 2020 International
  Joint Conference on Neural Networks (IJCNN). pp.~1--8 (2020).
  \doi{10.1109/IJCNN48605.2020.9206883}

\bibitem{ctftheory}
Petras, K., {ten Oever}, S., Jacobs, C., Goffaux, V.: Coarse-to-fine
  information integration in human vision. NeuroImage  \textbf{186},  103--112
  (2019). \doi{https://doi.org/10.1016/j.neuroimage.2018.10.086}

\bibitem{rtpowerline}
Pu, S., Xie, L., Ji, M., Zhao, Y., Liu, W., Wang, L., Yang, F., Qiu, D.:
  Real-time powerline corridor inspection by edge computing of uav lidar data.
  ISPRS - International Archives of the Photogrammetry, Remote Sensing and
  Spatial Information Sciences  \textbf{XLII-2/W13},  547--551 (06 2019).
  \doi{10.5194/isprs-archives-XLII-2-W13-547-2019}

\bibitem{pointnet}
Qi, C.R., Su, H., Mo, K., Guibas, L.J.: Pointnet: Deep learning on point sets
  for 3d classification and segmentation. CoRR  \textbf{abs/1612.00593} (2016),
  \url{http://arxiv.org/abs/1612.00593}

\bibitem{pointnetplusplus}
Qi, C.R., Yi, L., Su, H., Guibas, L.J.: Pointnet++: Deep hierarchical feature
  learning on point sets in a metric space. CoRR  \textbf{abs/1706.02413}
  (2017), \url{http://arxiv.org/abs/1706.02413}

\bibitem{2DUnet}
Ronneberger, O., Fischer, P., Brox, T.: U-net: Convolutional networks for
  biomedical image segmentation. CoRR  \textbf{abs/1505.04597} (2015),
  \url{http://arxiv.org/abs/1505.04597}

\bibitem{image2Object}
Rosman, B., Ramamoorthy, S.: Learning spatial relationships between objects. I.
  J. Robotic Res.  \textbf{30},  1328--1342 (10 2011).
  \doi{10.1177/0278364911408155}

\bibitem{kpconv}
Thomas, H., Qi, C.R., Deschaud, J., Marcotegui, B., Goulette, F., Guibas, L.J.:
  Kpconv: Flexible and deformable convolution for point clouds. CoRR
  \textbf{abs/1904.08889} (2019), \url{http://arxiv.org/abs/1904.08889}

\bibitem{SVMPowerline}
Wang, Y., Chen, Q., Liu, L., Zheng, D., Li, C., Li, K.: Supervised
  classification of power lines from airborne lidar data in urban areas. Remote
  Sensing  \textbf{9}(8) (2017). \doi{10.3390/rs9080771},
  \url{https://www.mdpi.com/2072-4292/9/8/771}

\bibitem{deeplearningpasedPowerline}
Yang, J., Huang, Z., Huang, M., Zeng, X., Li, D., Zhang, Y.: Power line
  corridor lidar point cloud segmentation using convolutional neural network.
  In: Lin, Z., Wang, L., Yang, J., Shi, G., Tan, T., Zheng, N., Chen, X.,
  Zhang, Y. (eds.) Pattern Recognition and Computer Vision. pp. 160--171.
  Springer International Publishing, Cham (2019)

\bibitem{rs13204029}
Zhao, J., Wang, Y., Cao, Y., Guo, M., Huang, X., Zhang, R., Dou, X., Niu, X.,
  Cui, Y., Wang, J.: The fusion strategy of 2d and 3d information based on deep
  learning: A review. Remote Sensing  \textbf{13}(20) (2021).
  \doi{10.3390/rs13204029}, \url{https://www.mdpi.com/2072-4292/13/20/4029}

\bibitem{isprs-annals-IV-2-W7-227-2019}
Zhou, M., Li, K.Y., Wang, J.H., Li, C.R., Teng, G.E., Ma, L., Wu, H.H., Li, W.,
  Zhang, H.J., Chen, J.Y., Chen, L.S.: Automatic extraction of power lines from
  uav lidar point clouds using a novel spatial feature. ISPRS Annals of the
  Photogrammetry, Remote Sensing and Spatial Information Sciences
  \textbf{IV-2/W7},  227--234 (2019).
  \doi{10.5194/isprs-annals-IV-2-W7-227-2019},
  \url{https://www.isprs-ann-photogramm-remote-sens-spatial-inf-sci.net/IV-2-W7/227/2019/}

\bibitem{dublincity}
Zolanvari, S.M.I., Ruano, S., Rana, A., Cummins, A., da~Silva, R.E., Rahbar,
  M., Smolic, A.: Dublincity: Annotated lidar point cloud and its applications.
  CoRR  \textbf{abs/1909.03613} (2019), \url{http://arxiv.org/abs/1909.03613}

\end{thebibliography}
\nocite{*}
\end{document}